\documentclass[11pt]{article}

\usepackage{emnlp2021}

\usepackage{times}
\usepackage{latexsym}

\usepackage[T1]{fontenc}

\usepackage[utf8]{inputenc}

\usepackage{microtype}
\usepackage{times}
\usepackage{latexsym}
\usepackage{microtype}
\usepackage{amsmath}
\usepackage{graphicx}
\usepackage{enumitem}
\usepackage{multirow}
\usepackage{multicol}
\usepackage{hyperref}

\usepackage{fancybox}
\usepackage{amsfonts}
\usepackage{todonotes}

\newcommand*\samethanks[1][\value{footnote}]{\footnotemark[#1]}
\title{Identifying and Mitigating Gender Bias in Hyperbolic Word Embeddings}

\author{
Vaibhav Kumar\thanks{\hspace{1mm}Authors have contributed equally.}\hspace{3mm} Tenzin Singhay Bhotia\samethanks \hspace{3mm} Vaibhav Kumar\samethanks \\
Delhi Technological University, India \\ \texttt{\{kumar.vaibhav1o1, tenzinbhotia0, vaibhavk992\}@gmail.com}
\AND
Tanmoy Chakraborty\\ Indraprastha Institute of Information Technology Delhi\\ \texttt{tanmoy@iiitd.ac.in}}

\begin{document}

\setlength{\abovedisplayskip}{3.0pt}
\setlength{\belowdisplayskip}{3.0pt}

\maketitle

\begin{abstract}
Euclidean word embedding models such as GloVe and Word2Vec have been shown to reflect human-like gender biases. In this paper, we extend the study of gender bias to the recently popularized hyperbolic word embeddings. We propose \textbf{gyrocosine bias}, a novel measure for quantifying gender bias in hyperbolic word representations and observe a significant presence of gender bias. To address this problem, we propose {\bf Poincaré Gender Debias} (\textbf{\texttt{PGD}}), a novel debiasing procedure for hyperbolic word representations. Experiments on a suit of evaluation tests show that \texttt{PGD} effectively reduces bias while adding a minimal semantic offset.
\end{abstract}

\section{Introduction}

Word embeddings are often used for word representation in a wide variety of tasks. 
However,  they have been shown to acquire gender stereotypes from the data which they are trained on \cite{bolukbasi2016man,caliskan2017semantics}. For instance, \citet{bolukbasi2016man} showed that a Word2Vec \cite{mikolov2013efficient} model trained on Google news corpus\footnote{\url{https://code.google.com/archive/p/word2vec/}}, generates analogy such as ``man is to computer programmer as woman is to homemaker". Therefore, given their pervasive usage, a large number of debiasing methods \cite{bolukbasi2016man,kaneko2019gender,bordia2019identifying,zhao2018learning,maudslay2019s} have recently been proposed.

The existing literature on mitigating gender bias exclusively focuses on word vectors which are embedded in an Euclidean space. However recently, non-Euclidean space has also been investigated for embedding words. For instance, motivated by the latent hierarchical structure of words, \citet{tifrea2018poincar} generalize GloVe \cite{pennington2014glove} to the hyperbolic space. As a result, they obtain Poincaré GloVe embedding, which achieves superior performance w.r.t GloVe in analogy, hypernymy detection, and word similarity tasks simultaneously. Despite the performance gain achieved, we observe that the issue of gender bias persists, as indicated by the biased analogies shown in Table \ref{table:1}.

\begin{table}[t!]
\centering
\scalebox{0.9}{
\begin{tabular}{|l|l|} 
 \hline
\multicolumn{2}{|c|}{\textbf{Gender Biased Analogies}}\\
\hline man \( \rightarrow \) doctor & woman \( \rightarrow \) nurse \\
\hline male \( \rightarrow \) supervisor & female \( \rightarrow \) assistant \\
\hline man \( \rightarrow \) programmer & woman \( \rightarrow \) storyteller \\
\hline he \( \rightarrow \) surgeon & she \( \rightarrow \) nurse \\
\hline man \( \rightarrow \) chairman & woman \( \rightarrow \) secretory \\
\hline
\end{tabular}}

\caption{Gender biased analogies generated by Poincaré GloVe \cite{tifrea2018poincar} trained on the English Wikipedia dump containing 1.4 billion tokens.}
\label{table:1}
\end{table}

In order to address this issue, we extend the study of gender bias to the hyperbolic space. We introduce {\bf gyrocosine bias}, a metric for quantifying the differential association of a given gender neutral word with a set of male and female definitional words using {\em gyrovector-based} \cite{ungar2008gyrovector} {\em formalism}. Further, we propose {\bf Poincaré Gender Debias} (\texttt{PGD}), a post-processing method for debiasing words embedded on the hyperbolic space. Experiments on a suite of evaluation tests show that \texttt{PGD} successfully mitigates gender bias while introducing minimal semantic and syntactic offset. {\em To the best of our knowledge, this is the first work that addresses the issue of gender bias in hyperbolic word embeddings.}


\section{Preliminaries}\label{prelim}
\paragraph{Hyperbolic Space and Poincaré Ball.}
An $n$-dimensional Hyperbolic space $\mathbb{H}^n$ is a non-Euclidean space with a constant negative curvature. Since a hyperbolic space cannot be embedded into an Euclidean space ~\cite{krioukov2010hyperbolic}, several isometric models have been proposed. 
Following \citet{tifrea2018poincar}, we use the Poincaré ball model of hyperbolic space. Formally, a Poincaré ball  $\left(\mathbb{D}^{n}_{c}, g^{\mathbb{D}}\right) $ of a radius $1/\sqrt{c}, c>0$ is defined by an $n$-dimensional manifold: $$\mathbb{D}^{n}_{c}=\left\{x \in \mathbb{R}^{n}:c\|x\|<1\right\}$$ having the Riemannian metric 
$g_{x}^{\mathbb{D}}$ as $
g_{x}^{\mathbb{D}}=\lambda_{x}^{2} g^{E}$, where $\lambda_{\mathbf{x}}=2 /\left(1-c\|\mathbf{x}\|^{2}\right)$ is the conformal factor, and $g^{E}=\mathbf{I}_{n}$ is the Euclidean metric tensor.

\paragraph{Gyrovector Space.}
Gyrovector space provides the algebraic setting for hyperbolic geometry (analogous to vector space for Euclidean geometry) \cite{ungar2008gyrovector}. We use gyrovector operations to quantify the bias in hyperbolic space. A brief description of the concepts used is given below\footnote{For a complete discussion, refer to \cite{ungar2008gyrovector}.}.

\paragraph{Möbius Addition and Subtraction.}{
The Möbius addition for $\space x, y \in \mathbb{D}^{n}_{c}$ is defined as:
\begin{equation*}
\resizebox{.8\hsize}{!}{$x \oplus_{c} y=\frac{\left(1+2 c\langle x, y\rangle+c\|y\|^{2}\right) x+\left(1-c\|x\|^{2}\right) y}{1+2 c\langle x, y\rangle+c^{2}\|x\|^{2}\|y\|^{2}}$}
\end{equation*}
where $\|\cdot\|$ denotes the Euclidean norm, and $\langle\cdot, \cdot\rangle$ is the Euclidean dot product. The Möbius subtraction $\ominus$ is then defined as $
a \ominus z=a \oplus(-z)$.}  
Note that in all cases, $c>0$ can be reduced to $c=1$ without any loss of generality. We refer to $\oplus_{1}$ as $\oplus$.

\paragraph{Rooted Gyrovectors.}
Any two ordered points $x,y \in \mathbb{D}^{n}$ give rise to a unique rooted gyrovector $v = \ominus x\oplus y$, which has its tail as $x$ and head as $y$. Additionally, any  $z \in \mathbb{D}^{n}$ can be identified as a rooted gyrovector with its tail at origin $O$ and head at $z$, which we represent as: $z' = O \oplus z$.

\paragraph{Gyrocosine Function.}
The gyrocosine is a measure of the gyroangle $\alpha$ ($0 \leq \alpha \leq \pi$) between two non-zero rooted gyrovectors $\ominus{A_{1}}\oplus{B_{1}}$ and $\ominus{A_{2}}\oplus{B_{2}}$, as given by: 
\begin{equation*}
\resizebox{.7\hsize}{!}{$
\cos \alpha=\frac{\ominus A_{1} \oplus B_{1}}{\|\ominus A_{1} \oplus B_{1}\|} \cdot \frac{\ominus A_{2} \oplus B_{2}}{\|\ominus A_{2} \oplus B_{2}\|}$}
\end{equation*}
\vspace{-1em}
\paragraph{Riemannian Optimization.}{\label{sec:optimize}
Let $f:\mathcal{M}\rightarrow \mathbb{R}$ be a smooth function to be optimized over the Riemannian manifold ($\mathcal{M}, p$). The Riemannian stochastic gradient descent update ~\cite{bonnabel2013stochastic} is then defined as: 
$$x_{t+1} \leftarrow exp_{x_t}(-\alpha g_t)$$ where $g_t \in T_{x_t}\mathcal{M}$ is the Riemannian gradient of $f_t$ at $x_t \in \mathcal{M}$, and $\alpha > 0$ is the learning rate. 
In this work, we use the Riemannian ADAM optimizer ~\cite{becigneul2018riemannian}\footnote{We use the implementations provided by the open source library: geoopt ~\cite{geoopt2020kochurov}}.}

\section{Gender Bias and its Mitigation}
\subsection{Bias in Hyperbolic Word Embeddings}

In this section, we first propose a pair of rooted gyrovectors to capture the differential gender information in the embedding space and then use them to quantify bias for any arbitrary word vector.

\paragraph{Gender Gyrovectors.} Given a set of words $\mathbb{W}$ embedded in an $n$-dimensional Poincaré ball $\mathbb{D}^{n}$, the gender gyrovectors are defined as follows:
\begin{align*}
    {g}_{mf} = \ominus \mu_M \oplus \mu_F \\ \newline
    {g}_{fm} = \ominus \mu_F \oplus \mu_M 
\end{align*}
where, $\mu_F$ and $\mu_M$ denote the intrinsic mean \cite{karcher1977riemannian, frechet1948elements} of a set of male definitional and female definitional hyperbolic word representations\footnote{We use the same set of gender definitional words as \cite{bolukbasi2016man}.} respectively (see Appendix 1.1 for details). 
We propose two gyrovectors for capturing the gender information because of the non-commutative property of the binary Möbius addition operator. 

\begin{figure}[!t]
\centering
\includegraphics[width=0.47\textwidth]{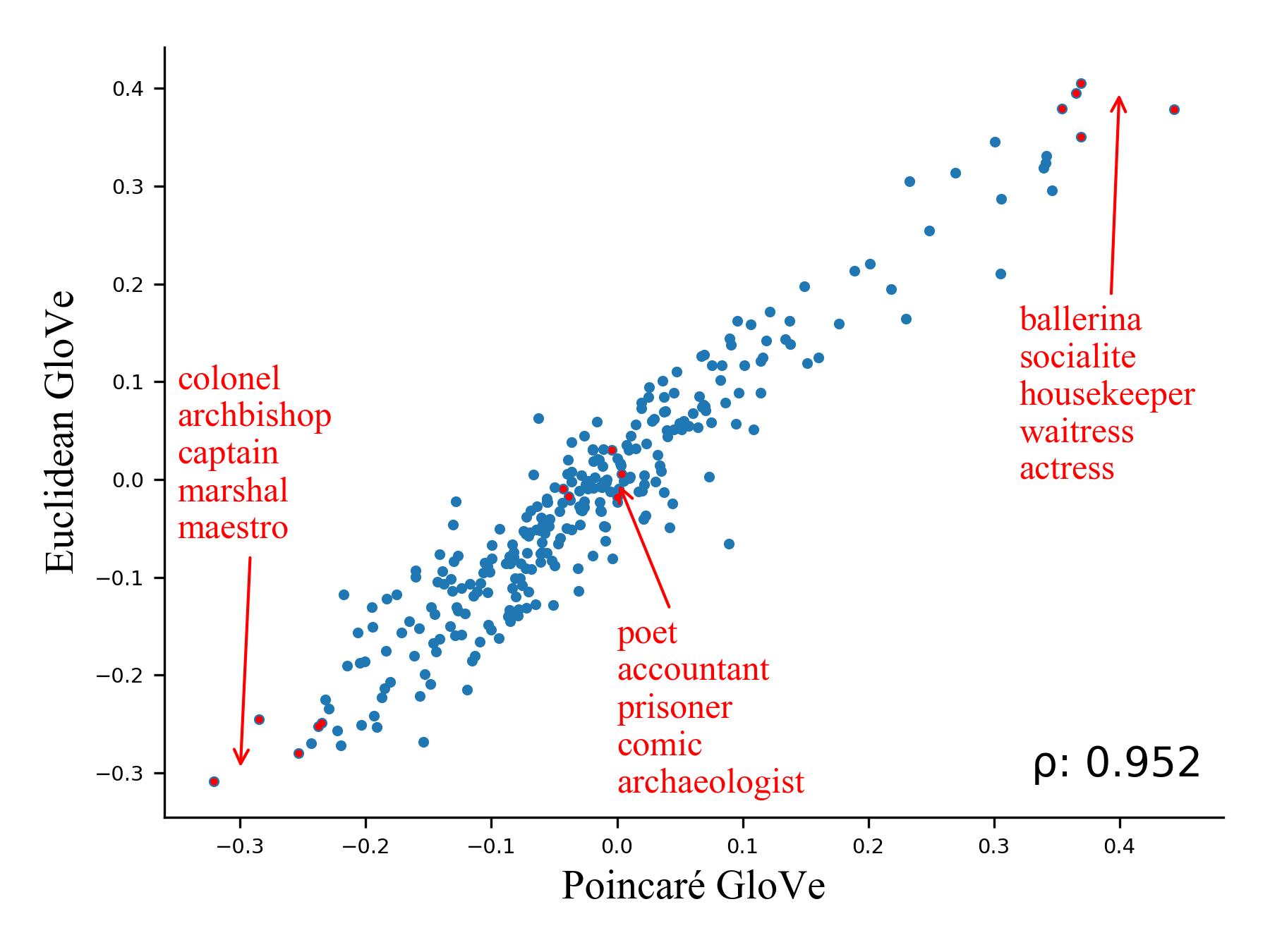}
\vspace{-2mm}
\caption{ Comparison of gender bias across GloVe and Poincaré GloVe (vectors of size $100D$ and trained on the English Wikipedia dump); \textit{x}-axis denotes gyrocosine bias, and \textit{y}-axis denotes direct bias. Each point in the scatter plot indicates a profession-related word.}
\label{corr}
\end{figure}

\paragraph{Gyrocosine Bias.} For a gender-neutral word $w$ embedded in an $n$-dimensional Poincaré ball $\mathbb{D}^{n}$, the gyrocosine bias is defined as follows:
\begin{equation*}
    \gamma(w) = 
\frac{\cos({{w'}}, {{g}_{mf}})-\cos({{w'}}, {{g}_{fm}})}{2}
\end{equation*}
where $\cos(.)$ is the gyrocosine function and $w'$ is the rooted gyrovector for $w$ as explained in Section \ref{prelim}.
Here, a positive value of $\gamma(w)$ implies that $w$ is female biased, while a negative value indicates male bias. To validate our metric w.r.t its Euclidean counterpart, we compare the gender bias for a set of profession words~\footnote{We use the list of professions provided by \citet{bolukbasi2016man}.} across GloVe and Poincaré GloVe in Figure \ref{corr}, using direct bias \cite{bolukbasi2016man} and gyrocosine bias respectively. We obtain a high Pearson correlation ($\rho = 0.952$) which indicates the competence of our metric in quantifying gender bias.


\subsection{Proposed Bias Mitigation Method}
We propose Poincaré Gender Debias (\texttt{PGD}), a multi-objective optimization based method for reducing gyrocosine bias in hyperbolic word embeddings while keeping the embeddings usable. To obtain the debiased counterpart ${w_{d}}$ of a given word ${w}$, \texttt{PGD} solves the following multi-objective optimization problem:
$$argmin_{{w'_d}}\big(F_s({w'_d}), F_g({w'_d})\big)$$
We use the weighted sum method to solve this optimization problem. More formally, we minimize the single objective function $F({w'_d})$ which can be expressed as follows:
\vspace{-4mm}
\begin{equation}
\begin{split}
    &F({w'_d}) \ = \ \lambda_{1}.F_{s}({w'_d}) +\lambda_{2}.F_{g}({w'_d})\\
&\text{such that\ }   \lambda_i \in [0, 1] \: \text{and} \:  \sum _i \lambda_i=1    
\end{split}
\label{obje}
\vspace{-4mm}
\end{equation}
$F({{w'_d}})$ is minimized using Riemannian optimization as described in Section ~\ref{sec:optimize}. We use the learning rate of  $3\times10^{-4}$ and equal objective weights, $\lambda_1=0.5$ and $\lambda_2=0.5$.
We now explain each component of Equation \ref{obje}; the range of each can be observed to be $[0,1]$.
\paragraph{$\boldsymbol{F_g({w'_d})}$:} This objective function aims to equalize word gyrovectors with the notion of male and female gender in the embedding space. It does so by minimizing the absolute difference of the gyrocosine values of the given word gyrovector and the two gender gyrovectors. The objective function is defined as follows:
\begin{align*}
    F_{g}({w}'_d) = \left | \cos({w'_d}, {g}_{mf})-\cos({w'_d}, {g}_{fm}) \right |/2
\end{align*}

\paragraph{$\boldsymbol{F_s({w'_d})}$:} This objective function aims to minimize the semantic offset caused by the debiasing procedure by maximizing the gyrocosine value between the rooted gyrovectors for a given word and its debias counterpart, thereby ensuring that the gyroangle between the two is minimum. The objective is defined as follows:
\begin{align*}
    F_{s}({w}'_d) &= |\cos({w}'_{d},{w}') -\cos({w}',{w}')|/2\\
    &= |\cos({w}'_{d},{w}') - 1|/2
\end{align*}

\begin{table*}[t]
    \centering
    \scalebox{0.65}
    {
        \begin{tabular}{|c|c c c c c c | c c c|}
            \hline
             \multirow{3}{*}{\bf Embedding} &  \multicolumn{6}{c|}{\bf (a) WEAT} &  \multicolumn{3}{c|}{\bf (b) SemBias}\\ \cline{2-10}
              & \multicolumn{2}{c|}{\textbf{$B_{1}$: Career vs Family}} & \multicolumn{2}{c|}{\textbf{$B_{2}$:  Math vs Art}} & \multicolumn{2}{c|}{\textbf{$B_{3}$:  Science vs Art}} & \textbf{Def} $\uparrow$ &  \textbf{Ster} $\downarrow$ & \textbf{None} $\downarrow$ \\
               &$p \uparrow$   &  \multicolumn{1}{c|}{$d \downarrow$} &        $p \uparrow$ & \multicolumn{1}{c|}{$d \downarrow$}           &        $p \uparrow$ & $d \downarrow$  & & &         \\
              \hline
             E-Glove  &   0.0773 & \multicolumn{1}{c|}{0.7423} & 0.4186 & \multicolumn{1}{c|}{0.1716} & 0.5443 & -0.1131 & 79.5 & 14.8 & \bf 5.7 \\
             P-GloVe & 0.0418 & \multicolumn{1}{c|}{0.8929} & 0.3045 & \multicolumn{1}{c|}{0.2929} & 0.2138 & 0.4568 & 75.3 & 7.0 & 17.7 \\
             \hline
             {\tt PGD}-GloVe & \bf 0.2827 & \multicolumn{1}{c|}{\bf 0.3054} & \bf 0.8292 &  \multicolumn{1}{c|}{\bf -0.5379} & \bf 0.6015 & \bf -0.1984& \bf 83.0 & \bf 0.0 & 17.0\\
             \hline
        \end{tabular}
    }
    
\caption{Comparison of different embedding models on (a) WEAT for different sets of target words - $B_{i}$ and (b) SemBias dataset. $\uparrow (\downarrow)$ indicates that higher (lower) value is better. In (a), a $p$-value $>$ 0.5 denotes insignificant bias, and a lower Cohen's $d$ shows lesser bias. We use (negative) Poincaré distance as a measure of similarity for P-GloVe and \texttt{PGD}-GloVe (see Appendix 2.1 for details).}
    \label{tab:wese}
    \vspace{1em}
\end{table*}

\begin{table*}[t]
    \centering
    
    \scalebox{0.65}{
    
    \begin{tabular}{|c|c c c c c c | c c c|}
        \hline
         \multirow{2}{*}{\bf Embedding} &  \multicolumn{6}{c|}{\bf (a) Semantic} &  \multicolumn{3}{c|}{\bf (b) Analogy}\\
         & RW & WordSim &  SimLex & SimVerb & MC & RG & Google-Sem & Google-Syn & MSR \\
         \hline
         E-Glove & 0.3787 & 0.5668 & 0.2964 & 0.1639 & 0.6562 & 0.6757 & 0.6427 & 0.5950 & 0.4826 \\
         P-GloVe & \bf 0.4190 & \bf 0.6207 & \bf 0.3210 & \bf 0.1915 & \bf 0.7883 & \bf 0.7596 & 0.6629 & \bf 0.6086 & \bf 0.4958 \\
         \hline
         {\tt PGD}-GloVe       & 0.4170 & 0.6152 & 0.3215 & 0.1914 & 0.7873 &  0.7569 & \bf 0.6635 & 0.6069 & 0.4938\\
         \hline
    \end{tabular}
    }
    \caption{Comparison of different embedding models in (a) semantic and (b) analogy tests. The performance of {\tt PGD}-GloVe is at par with other baselines, indicating the addition of minimal semantic disturbance.}
    \label{tab:Sem_anl}
\end{table*}
\vspace{-1em}
\section{Experiments and Results}
\label{sec:length}
\subsection{Dataset and Baselines}
We use the following pre-trained word embedding models provided by \citet{tifrea2018poincar} as our baseline, each of which is trained on the English Wikipedia dump and consists of 189,533 tokens: $100D$ Poincaré GloVe, \( h(x)=\cosh ^{2}(x) \), with init trick\footnote{Refer \citet{tifrea2018poincar}.} -- referred to as \textbf{P-GloVe};  $100D$ Euclidean GloVe, with init trick -- referred to as \textbf{E-Glove}. We debias Poincaré GloVe using \texttt{PGD} to obtain \textbf{\texttt{PGD}- GloVe}. Similiar to  \citet{bolukbasi2016man}, we split the vocabulary into a set of gender specific $S$ and gender neutral words  $N$. Only $N$, consisting of $126,065$ words, is debiased.



\subsection{Evaluation Metrics and Results}

\paragraph{Word Embedding Association Test (WEAT):}{
\citet{caliskan2017semantics} introduce WEAT, a hypothesis testing method for analyzing stereotypical biases present in word embeddings. WEAT comprises two sets of equal-sized target words -- $X$ (like `engineer', `warrior')  and $Y$ (like  `nurse', `receptionist'), and two sets of  equal sized attribute words -- $A$ (like `he', `male') and $B$ (like `she', `female')~\footnote{Refer Appendix 2.2 for the complete list of words.}. The aim of the test is to determine if the set $X$ or $Y$ is more biased towards one gender than the other.
We report Cohen’s $d$ and $p$-value for three categories of target words in Table \ref{tab:wese}(a). Evidently, \texttt{PGD}-GloVe achieves consistent performance for all sets of target words, reflecting the efficient reduction in bias (see Appendix 2.2 for more details).
Further, we observe that Poincaré GloVe is more biased than Euclidean GloVe, having a lower $p$-value and a higher Cohen's $d$ for all the target word categories. Since, the training data and variable configurations like the dimensions and initialization were same, a higher bias could be due to the negative curvature of hyperbolic space and would remain as an interesting future direction.
 }
    
    
\paragraph{SemBias: }{

SemBias \cite{zhao2018learning} is a word analogy dataset used for measuring the ability of word embeddings to generate biased analogies. Each instance of SemBias consists of four types of word pairs -- gender-definition word pair (\textbf{Def}; e.g., ‘wizard-witch’), gender-stereotype word pair (\textbf{Ster};  e.g., ‘doctor-nurse’) and two other word pairs having similar meanings but no gender-based relation (\textbf{None}; e.g., ‘salt-pepper’). For each instance, we choose the most probable analogy estimated by the embedding model. The analogy computation for word embeddings in Poincaré ball model uses a hyper-parameter~\footnote{Refer Appendix 2.3 for more details.} $t \in [0,1]$ \cite{tifrea2018poincar}. Following \citet{levy2015improving, tifrea2018poincar}, we use 2-fold cross-validation over the gender-definition word pair analogies to estimate the value of $t$ as $0.3$. From Table \ref{tab:wese}(b), we can observe that \textbf{Ster} analogies are completely inhibited for \texttt{PGD}-GloVe along with a 10.2 \% relative improvement in \textbf{Def} w.r.t P-GloVe. This implies that \texttt{PGD} dramatically mitigates the preference of a gender-neutral word between male and female words, thereby improving its efficacy to form correct analogies. We further observe that P-GloVe has a lesser ($\textbf{Ster}$) score as compared to E-GloVe. However, this result is attained by a lower performance in $\textbf{Def}$ and $\textbf{None}$, indicating the superiority of E-GloVe.

}        


%
\paragraph{Semantic Tests: }
Word semantic similarity tests determine the correlation between the similarity of word pairs obtained through the word embeddings and the similarity ratings given by humans.
We report Spearman's correlations for the following benchmark datasets: RW \cite{sim-RW}, WordSim \cite{sim-WordSim}, SimLex \cite{sim-Simlex}, SimVerb \cite{sim-SimVerb}, MC \cite{sim-MC}, and RG \cite{sim-RG}. 
Following \citet{tifrea2018poincar}, we use the negative Poincaré distance as the similarity metric for P-GloVe and {\tt PGD}-GloVe while for E-GloVe, we use the regular cosine similarity. 
Table \ref{tab:Sem_anl}(a) shows that {\tt PGD}-GloVe successfully retains most of semantic information. Therefore, \texttt{PGD} produces minimal semantic offset.

\paragraph{Analogy Tests: }
Analogy tests assess the efficacy of a word embedding model to answer the following question : `$w_1$ is to $w_2$ as $w_3$ is to ?'. Here, unknown word $w_4$ is estimated as the closest word vector to the arithmetic: $w_2 - w_1 + w_3$. Following \citet{tifrea2018poincar}, we estimate the closest word vector $w_4$ using gyro-translations and the cosine similarity metric for the following benchmarks: Google Sem, Syn \cite{analogy-Google} and MSR \cite{analogy-MSR}. Table \ref{tab:Sem_anl}(b) shows that {\tt PGD}-GloVe closely follows P-GloVe in all the analogy measures. The offsets are so minimal that {\tt PGD}-GloVe retains the semantic and analogy advantages of P-GloVe over E-GloVe.

\section{Conclusion}
In this paper, we presented the first study of gender bias in a non-Euclidean space. We proposed {\bf gyrocosine bias},  a metric for quantifying gender bias in hyperbolic embedding. We observed that stereotypical gender biases permeate the Poincaré GloVe model. We also proposed {\bf Poincaré Gender Debias}, a post-processing method for debiasing hyperbolic word embeddings. Experimental results indicated that our method successfully minimizes gender bias while retaining the practical usability of  word embeddings.

\bibliography{custom.bib}
\bibliographystyle{acl_natbib}

\end{document}


\maketitle

\begin{table*}[t]
    \centering
    \scalebox{0.8}{
    \begin{tabular}{|c|l|c |l|}
    \hline
        & \multirow{2}{*}{\bf Attribute Words} & M & brother, father, uncle, grandfather, son, he, his, him\\

        & & F & sister, mother, aunt, grandmother, daughter, she, hers, her\\
        \hline
         & \multirow{2}{*}{$B_{1}$ : Career vs Family} & X & executive, management, professional, corporation, salary, office, business, career\\
        \multirow{2}{*}{\rotatebox{90}{\bf Target Words}} & & Y & home, parents, children, family, cousins, marriage, wedding, relatives\\
        \cline{2-4}
        & \multirow{2}{*}{$B_{2}$ : Maths vs Arts} & X & math, algebra, geometry, calculus, equations, computation, numbers, addition\\
        & & Y & poetry, art, Shakespeare, dance, literature, novel, symphony, drama\\
        \cline{2-4}
        & \multirow{2}{*}{$B_{3}$ : Science vs Arts} & X & science, technology, physics, chemistry, Einstein, NASA, experiment, astronomy\\
        & & Y & poetry, art, Shakespeare, dance, literature, novel, symphony, drama\\
        \hline
    \end{tabular}
    }
    \caption{List of attribute and target words for each category used for our experiments.}
    \label{tab:weat_data}
    \vspace{1em}
\end{table*}

\begin{table*}[t]
    \centering
    \scalebox{0.8}{
    \begin{tabular}{|c|c c|c c|c c|}
        \hline
         \multirow{2}{*}{\bf Embedding} &  \multicolumn{2}{c|}{\textbf{$B_{1}$: Career vs Family}} & \multicolumn{2}{c|}{\textbf{$B_{2}$:  Math vs Art}} & \multicolumn{2}{c|}{\textbf{$B_{3}$:  Science vs Art}} \\
                         &      $p$   &  $d$ &        $p$ & $d$           &        $p$ & $d$            \\
        \hline
        Euclidean GloVe  &   0.0773 & 0.7423 & 0.4186 & 0.1716 & 0.5443 & -0.1131 \\ \hline
        Poincaré  GloVe (cosdist)  &   0.0329 & 0.9423 & 0.2422 & 0.4169 & 0.1733 & 0.5267 \\
        Poincaré  GloVe (-pdistance)  &   0.0418 & 0.8929 & 0.3045 & 0.2929 & 0.2138 & 0.4568 \\
        \hline
        {\tt PGD}  GloVe (cosdist)  &   0.2456 & 0.3628 & \textbf{0.8759} & \textbf{-0.6310} & 0.5830 & -0.1559\\
        {\tt PGD}  GloVe (-pdistance)  &   \textbf{0.2827} & \textbf{0.3054} & 0.8292 & -0.5379 & \textbf{0.6015} & \textbf{-0.1984}\\
        \hline
    \end{tabular}
    }
    \caption{Compairison of word embedding on WEAT for different categories - $B_{i}$. A higher $p$ and lower $d$ is better. Cos dist represents the use of cosine similarity as the similarity metric while -pdistance represents the (negative) Poincaré distance}
    \label{tab:weat}
\end{table*}

\section{Poincare Gender Debiasing}
\subsection{Intrinsic Mean Calculation}
To compute gender direction gyrovectors, we calculate the the intrinsic means $\mu_F$ and $\mu_M$ for a set of male and female definitional words which are obtained from the list provided by \citet{bolukbasi2016man}~\footnote{\url{https://github.com/tolga-b/debiaswe}}. We use the following definition of intrinsic mean ~\cite{karcher1977riemannian, frechet1948elements} for points in a Riemannian Manifold.
\begin{mydef}
Let ($\mathcal{M}, d$) be a Riemannian Manifold then the intrinsic mean $\mu$ for a set of points $x_1,...,x_n \in \mathcal{M}$ is defined as the minimizer of the squared sum ditances to each point
\begin{equation}\label{eq:mean}
    \mu = argmin_{x\in \mathcal{M}}\sum_{i=1}^n d(x, x_i)^2
\end{equation}
\end{mydef}
The Equation \ref{eq:mean} is optimized using the Riemannian ADAM optimizer ~\cite{becigneul2018riemannian} provided by the geoopt python library ~\cite{geoopt2020kochurov}. To speed up the optimization process, we initialize $\mu$ with the arithmetic mean of the given points and use a learning rate of 0.0003.

\subsection{PGD Details}
We use the pre-trained Poincaré GloVe embeddings provided by \citet{tifrea2018poincar}~\footnote{\url{https://github.com/alex-tifrea/poincare_glove}}. For \texttt{PGD}, we optimize the proposed objective function using the Riemannain ADAM optimizer ~\cite{becigneul2018riemannian} following the implementations provided by the geoopt ~\cite{geoopt2020kochurov}~\footnote{\url{https://github.com/geoopt/geoopt/}} python library.  We set the objective function weights equal to each other ($\lambda_1=0.5$ and $\lambda_2=0.5$). The learning rate of $\alpha=0.0003$ is used and we run the PGD optimization process for 350 epochs.





\section{Results}
\subsection{Poincaré Distance}
We use (negative) Poincaré distance to evaluate similarity of vectors in analogy tasks. The Poincaré distance between two points $\space x, y \in \mathbb{D}^{n}_{c}$ is defined using the following expression:
$$d_{\mathbb{D}}(\mathbf{x}, \mathbf{y})=\frac{2}{\sqrt{c}} \tanh ^{-1}\left(\sqrt{c}\left\|-\mathbf{x} \oplus_{c} \mathbf{y}\right\|\right)$$
where, $\oplus_{c}$ denotes the \textit{Möbius Addition} operation. Without any loss of generality we use $c = 1$. 

\subsection{WEAT}

A description of the different set of words ~\cite{chaloner2019measuring} used for conducting WEAT is given in Table \ref{tab:weat_data}. For \texttt{PGD} GloVE and Poincaré Glove, we can use both negative Poincaré distance (-pdistance) and cosine distance as a measure of similarity, Table \ref{tab:weat} shows a comparison using both of them. Table \ref{tab:weat} further corroborates the debiasing efficacy of \texttt{PGD}.

\subsection{Analogy tests}
The aim of analogy tests is to find a word $d$, such that $d$ is to a word $c$, as the word $b$ is to the word $a$. In Euclidean space, $d = c+(b-a) = b+(c-a)$. However, in the hyperbolic space, because of the non-zero curvature, there are two possible solutions to $d$, created using gyro-translations as follows ~\cite{tifrea2018poincar}:
$$d_1 = c \oplus gyr[c, \ominus a](\ominus a \oplus b) $$
$$d_2 = b \oplus gyr[b, \ominus a](\ominus a \oplus c) $$
Where $gyr$ is the gyro operator ~\cite{ungar2008gyrovector} defined as follows:
$$gyr[a,b]c = \ominus (a \oplus b) \oplus \{ a \oplus (b \oplus c)\}$$

We combine the two solutions into a single point, $d^t_{d_1d_2}$, using a hyper-parameter $t \in [0,1]$ as follows:
$$d^t_{d_1d_2} = d_1 \oplus((-d_1 \oplus d_2) \otimes t) $$
Here, $t$ is a hyper-parameter, $\oplus$ is the Möbius addition operator as described in the manuscript while $\otimes$ is the Möbius scalar multiplication defined as follows:
\begin{mydef}
Möbius Scalar Multiplication of a $x \in \mathbb{D}^n \setminus \left \{ \bf{0} \right \}$ by a real number $r \in \mathbb{R}$ is defined as
$$r \otimes x = tanh(rtanh^{-1}(\left \| x \right \|)) \frac{x}{\left \| x \right \|} $$
\end{mydef}
We follow the same approach as \citet{tifrea2018poincar} for estimating $t$, using 2-fold cross-validation and vary $t$ across 11 values in $\{0, 0.1, ..., 1\}$ to find the best value corresponding to highest analogy score. 

\bibliography{custom.bib}
\bibliographystyle{acl_natbib}